\documentclass[sigconf,nonacm]{acmart}

\usepackage[capitalise]{cleveref}
\usepackage{booktabs}

\usepackage{pifont}
\newcommand{\xmark}{\ding{55}}

\usepackage{makecell} 

\def\CPP{C\raisebox{0.5ex}{\tiny\textbf{++}}}

\setcopyright{none}

\begin{document}

\title{Call for Action: Towards the Next Generation of Symbolic Regression Benchmark}

\author{Guilherme Seidyo Imai Aldeia}
\email{guilherme.aldeia@ufabc.edu.br}
\orcid{0000-0002-0102-4958}
\affiliation{%
  \institution{Federal University of ABC}
  \city{Santo André}
  \state{São Paulo}
  \country{BR}
}

\author{Hengzhe Zhang}
\orcid{0000-0002-2254-8304}
\email{hengzhe.zhang@ecs.vuw.ac.nz}
\affiliation{%
  \institution{Victoria University of Wellington}
  \city{Wellington}
  \country{NZ}}

\author{Geoffrey Bomarito}
\orcid{0000-0002-5540-6871}
\email{geoffrey.f.bomarito@nasa.gov}
\affiliation{%
 \institution{NASA Langley Research Center}
 \city{Hampton}
 \state{Virginia}
 \country{USA}}

\author{Miles Cranmer}
\orcid{0000-0002-6458-3423}
\email{mc2473@cam.ac.uk}
\affiliation{%
  \institution{University of Cambridge}
  \city{Cambridge}
  \country{UK}}

\author{Alcides Fonseca}
\email{me@alcidesfonseca.com}
\orcid{0000-0002-0879-4015}
\affiliation{%
  \institution{LASIGE, Faculdade de Ciências da Universidade de Lisboa}
  \city{Lisboa}
  \country{PT}}
  
\author{Bogdan Burlacu}
\orcid{0000-0001-8785-2959}
\affiliation{%
  \institution{University of Applied Sciences Upper Austria}
  \state{Upper Austria}
  \country{AT}}

\author{William G. La Cava}
\email{william.lacava@childrens.harvard.edu}
\orcid{0000-0002-1332-2960}
\affiliation{%
  \institution{\mbox{\hspace{-1.475em}Computational Health Informatics Program}\\Boston Children's Hospital\\Harvard Medical School}
  \city{Boston}
  \state{Massachusetts}
  \country{USA}}
  
\author{Fabrício Olivetti de França}
\orcid{0000-0002-2741-8736}
\email{folivetti@ufabc.edu.br}
\affiliation{%
  \institution{Federal University of ABC}
  \city{Santo André}
  \state{São Paulo}
  \country{BR} 
}

\renewcommand{\shortauthors}{Imai Aldeia et al.}

\begin{abstract}
Symbolic Regression (SR) is a powerful technique for discovering interpretable mathematical expressions. However, benchmarking SR methods remains challenging due to the diversity of algorithms, datasets, and evaluation criteria. In this work, we present an updated version of SRBench. Our benchmark expands the previous one by nearly doubling the number of evaluated methods, refining evaluation metrics, and using improved visualizations of the results to understand the performances. Additionally, we analyze trade-offs between model complexity, accuracy, and energy consumption. Our results show that no single algorithm dominates across all datasets.
We propose a \textit{call for action} from SR community in maintaining and evolving SRBench as a living benchmark that reflects the state-of-the-art in symbolic regression, by standardizing hyperparameter tuning, execution constraints, and computational resource allocation. We also propose deprecation criteria to maintain the benchmark's relevance and discuss best practices for improving SR algorithms, such as adaptive hyperparameter tuning and energy-efficient implementations.
\end{abstract}


  
\maketitle

\section{Introduction}

Symbolic regression (SR)~\cite{Koza1992,kronberger2024} is a supervised machine learning technique that searches for a mathematical expression $\hat{f}(\mathbf{x}, \mathbf{\hat{\theta}})$ that fits a set of points $\mathbf{x}_i \in \mathcal{X}, y_i \in \mathcal{Y}$ such as $y \approx \hat{f}(\mathbf{x}, \mathbf{\hat{\theta}})$, where $\mathbf{\hat{\theta}}$ is a vector of adjustable parameters.
Searching for the function $\hat{f}$ adds a degree-of-freedom to data fitting that differs from traditional parametric models that starts from a fixed $f(x, \mathbf{\theta})$ and tries to find $\mathbf{\hat{\theta}}$ that minimizes a loss function.
Currently, many SR algorithms implement a variation of the original genetic programming (GP) as proposed by Koza~\cite{Koza1992,koza1994genetic}.
Since then, it has been successfully applied to many real-world applications, such as, scientific discovery in engineering~\cite{LIU2024108986}, healthcare~\cite{LaCava2023}, physics~\cite{HRUSKA2025118821}, among others~\cite{schmidt2009distilling,kronberger2024,cranmer2023interpretablemachinelearningscience,udrescu2020ai,multiviewsr}.
Its application in physics can be seen as ``automating Kepler'' ---discovering analytical expressions through trial and error---, particularly evident in recent works that use SR to recover laws directly from experimental data~\cite{Makke2024}, underscoring its potential as a tool for automated scientific reasoning.

Equation discovery is an NP-hard problem~\cite{virgolin2022symbolicregressionnphard}, and implementations rely on various approaches to solve it, such as GP~\cite{Schmidt2010, la2016inference, randall2022bingo, eplex, cava2018learning, espada2022data, Virgolin_2021, 10.1145/3377930.3390237, itea, kommenda2020parameter, pstree, cranmer2023interpretablemachinelearningscience, tir}, Neural Networks~\cite{pmlr-v139-biggio21a,pmlrv80sahoo18a}, Transformers~\cite{shojaee2023transformerbased,NEURIPS2022_dbca58f3,kamienny2022endtoend}, Bayesian models~\cite{jin2020bayesiansymbolicregression}, iterative methods~\cite{mcconaghy2011ffx,Kartelj2023,de2018greedy}, and even exhaustive approaches~\cite{Kammerer_2020,Bartlett_2024}.
In its original implementation, GP relied on generating ephemeral random constants as part of the expression, which could lead to sub-optimal exploration of the search space if these constants were not properly set. This was alleviated with the use of linear scaling~\cite{keijzer2003improving} by scaling and translating the solution before measuring the loss function. Modern implementations rely on the optimization of the numerical parameters by either enforcing the creation of a model that is linear in the parameters~\cite{arnaldo2014multiple,itea,tir,cava2018learning} or applying a non-linear optimization method to fit the parameters~\cite{10.1145/3377930.3390237,kommenda2020parameter}.

Despite recent advances in the field of SR, \citet{la2021contemporary} observed a lack of consensus regarding a standardized set of datasets and benchmarking methodologies, making it challenging to accurately establish the current state-of-the-art (SotA). Another issue is the lack of a unified programming interface and clear instructions on installing and using each approach properly, creating barriers to new experimentation and reproducibility.
To address these challenges, SRBench~\footnote{\url{https://cavalab.org/srbench/}} was proposed as a living benchmark for SR, comparing different SR implementations and standard ML regression models. This benchmark enforced the adoption of a common Python API and installation scripts within a \emph{sandbox} environment, making it easier to install and evaluate different algorithms.

Running a benchmark is a laborious yet important task.
Not only is it necessary to compare methods, but it also serves as a tool for measuring progress in the field.
While the SRBench was a significant step toward comparing and understanding the current SotA in SR, designing effective benchmarks often requires further refinements and consideration of new factors to ensure their usefulness and longevity.
An effective benchmark should contain diverse subclasses of problems, identifying which algorithms perform best in each scenario, mapping problems to algorithms rather than providing a one-size-fits-all solution.
Additionally, the benchmark must challenge contemporary SR algorithms while remaining computationally feasible and representative of real-world challenges, ensuring that results extend beyond artificial test cases.
SRBench falls short in these aspects, as it relies on more than $200$ datasets without a detailed categorization of their specific challenges. Furthermore, it presents final results using only aggregated metrics, which can hide finer details in performance comparisons.

In this paper, we propose several improvements to the existing SRBench (hereafter referred to as SRBench 1.0), identifying challenges that must be addressed to develop a more refined version of the benchmark (SRBench 2.0).
Additionally, this paper serves as a \emph{call to action}, encouraging the community to engage in discussions on best practices for benchmarking and evaluating SR algorithms. Such involvement is crucial for the field to achieve a more mature state and drive further advancements.

The primary improvements proposed include: expanding the benchmark to incorporate new SR algorithms, bringing the total to $25$; increasing the number of independent runs from $10$ to $30$ to enhance the statistical significance of the results; selecting $24$ datasets, divided into two tracks: \textit{(i)} black-box (drawn from SRBench 1.0) and \textit{(ii)} phenomenological \& first-principles (sourced from~\cite{cranmer2023interpretablemachinelearningscience}); and refining the reporting methodology to provide a more detailed comparison of the nuances among top-performing algorithms, rather than simply ranking them and designating a single SotA method.

This \emph{call to action} aims to highlight the challenges and issues encountered during benchmarking and to propose features that could streamline the process. Addressing these issues requires collaboration from method's authors/maintainers to develop and extend a common API that facilitates the expansion of SRBench in future iterations.
To our knowledge, this is the largest study comparing SR methods under a unified experimental setup.

The paper is organized as follows:
\Cref{sec:related_work} reviews previously published benchmarks on symbolic regression, highlighting differences to our work.
\Cref{sec:methods} presents our proposed update to SRBench, with each subsection addressing a key aspect of our update to the benchmark.
\Cref{sec:results} reports our results, using different levels of aggregation to gain deeper insights into the performance of individual algorithms.
\Cref{sec:discussions} discusses the results, and \Cref{sec:conclusions} presents our conclusions and call to action for advancing the benchmarking of symbolic regression methods.

\section{Related work}~\label{sec:related_work}

In 2018, \citet{wherearewenow} surveyed many SR papers showing that methods were being benchmarked on simple problems with a lack of standardization, which could diminish the perception of SR achievements outside of its field. They also acknowledged previous efforts to conduct benchmarks for SR methods, then conducted an experiment with four SR algorithms comparing with different ML methods across more than $100$ datasets of up to $1000$ samples.

\citet{la2021contemporary} followed up the benchmark efforts and proposed SRBench, a joint-effort to achieve a large scale comparison of $14$ modern SR algorithms comprising GP-based, deterministic, and deep learning based approaches. In this paper, the authors mention that determining a state-of-the-art should not be the sole focus of research --- however, promising avenues of investigation cannot be well-informed without empirical evidence. In SRBench, the authors increased the number of problems and their dimensionalities, using datasets from PMLB~\cite{Olson2017PMLB, romano2021pmlb}, an open-source library for benchmarking ML methods. A new ground-truth track was also added with more than $100$ physics equations from Feynman lectures~\cite{feynman2006feynman,feynman2015feynman} with data generated synthetically. They  provided a more informative view of the current state of the field. Given the large number of problems and algorithms, they performed $10$ individual runs for each dataset, and the budget was based on evaluations --- which is hard to measure uniformly across the variety of different approaches with different computational runtime.

Further discussions emerged in top-tier conferences related to SR (such as GECCO), with criticisms about aggregated views of the results~\cite{dick2022genetic,10.1145/3597312}. 
\citet{srbenchplusplus} benchmarked a smaller selection of SR algorithms, including more recent approaches, on a proposed set of artificially generated datasets to measure the performance in specific tasks: overall accuracy performance, robustness to noise, capabilities of selecting the relevant features, extrapolation performance, and interpretability. With a smaller selection of algorithms and datasets it was possible to make precise statements of the current state of SR and the current weak points that should be addressed in the future. The authors also introduced a runtime limit instead of using the number of evaluations, stimulating the efficient implementation of the algorithms. The authors noticed that even though SR shares many shortcomings with traditional ML models, they have a potential to overcome it using domain knowledge and allowing a customized experience to the user~\cite{Ingelse2023}. Regarding the benchmarks, the authors feel that there is still room for improvements on how to craft datasets that correctly mimic the challenges faced in the real-world.

\citet{matsubara2024rethinking} noticed that the Feynman dataset~\cite{udrescu2020ai} used in the SRBench's ground-truth track did not adequately depict the original physical phenomena because of how the values were sampled --— the data generation was originally done by uniformly sampling standardized values. They proposed to replace the synthetic data generation process with normal or logarithmic distributions that mimic what we would empirically observe if we were to collect the data in practice, and evaluated six different SR methods.

\citet{egklitz2020} proposes a benchmark with open-source and ease-of-use principles. They evaluated four SR methods on five synthetic and four real-world problems.
\citet{esg1352} benchmarks only five SR methods focusing on benchmarking SR for learning equations in civil and construction engineering, where equations are traditionally derived using linear regression.

\citet{thing2025cp3} proposed an unified tool for benchmarking SR in cosmology, including $12$ SR methods and $28$ datasets representing cosmological and astroparticles physics problems, finding that most methods performed poorly in the benchmark. They argue that using standard datasets (\textit{i.e.} PMLB) is prudent for academic comparisons, but these ensembles of datasets do not provide a representative measure for specific problems. They conducted off-the-shelf evaluations, which are useful because they do not require extra effort from the user, and, to improve reproducibility, they package the $12$ methods into Docker containers.

\section{Methods}~\label{sec:methods}

\subsection{Dataset selection}

For this version of the benchmark, we propose two distinct tracks: \textit{black-box} and \textit{phenomenological \& first-principles}. \cref{tab:datasets} lists the datasets used in our benchmark, along with the number of samples (\# Rows), features (\# Cols), and the range of values for the target feature (codomain).
For the latter track, all codomains consist of real values, so instead we provide the data sources for these datasets.
All datasets contain no missing values, have only numeric features, and are standardized before training.

\begin{table}[htb]
    \caption{
        Metadata for each benchmark track. Dataset names match their corresponding PMLB entries
    }
    \label{tab:datasets}
    \footnotesize
    \begin{tabular}{@{}p{2.75cm}rrl@{}}
        \toprule\midrule
        \textbf{Black-box} & \textbf{\# Rows} & \textbf{\# Cols} & \textbf{Codomain} \\ \midrule
        1028\_SWD & 1000 & 11 & $\mathbb{Z}^{+}$ \\
        1089\_USCrime & 47 & 14 & $\mathbb{Z}^{+}$ \\
        1193\_BNG\_lowbwt & 31104 & 10 & $\mathbb{R}^{+}$ \\
        1199\_BNG\_echoMonths & 17496 & 10 & $\mathbb{R}$ \\
        192\_vineyard & 52 & 3 & $\mathbb{R}^{+}$ \\
        210\_cloud & 108 & 6 & $\mathbb{R}^{+}$ \\
        522\_pm10 & 500 & 8 & $\mathbb{R}^{+}$ \\
        557\_analcatdata\_apnea1 & 475 & 4 & $\mathbb{Z}^{+}$ \\
        579\_fri\_c0\_250\_5 & 250 & 6 & $\mathbb{R}$ \\
        606\_fri\_c2\_1000\_10 & 1000 & 11 & $\mathbb{R}$ \\
        650\_fri\_c0\_500\_50 & 500 & 51 & $\mathbb{R}$ \\
        678\_visualizing\_environmental & 111 & 4 & $\mathbb{R}^{+}$ \\ \midrule
        \makecell[l]{\textbf{Phenomenological \&}\\\textbf{first-principles}} & \textbf{\# Rows} & \textbf{\# Cols} & \textbf{Data source} \\ \midrule
        first\_principles\_absorption & 14 & 2 & \citet{multiviewsr} \\
        first\_principles\_bode & 8 & 2 & \citet{bonnet1764contemplation} \\
        first\_principles\_hubble & 32 & 2 & \citet{Hubble1929} \\
        first\_principles\_ideal\_gas & 30 & 4 & Generated, 10\% noise \\
        first\_principles\_kepler & 6 & 2 & \citet{keplerHarmonicesMundi1619} \\
        first\_principles\_leavitt & 26 & 2 & \citet{leavittPeriods25Variable1912} \\
        first\_principles\_newton & 30 & 4 & Generated, 10\% noise \\
        first\_principles\_planck & 100 & 3 & Generated, 10\% noise \\
        first\_principles\_rydberg & 50 & 3 & Generated, 1\% noise \\
        first\_principles\_schechter & 27 & 2 & Generated, 20\% noise \\
        first\_principles\_supernovae\_zr & 236 & 2 & \citet{multiviewsr} \\
        first\_principles\_tully\_fisher & 18 & 2 & \citet{tullyNewMethodDetermining1977} \\ \midrule\bottomrule
    \end{tabular}
\end{table}

The black-box track is a selection of $12$ regression datasets from the PMLB 1.0 project~\cite{romano2021pmlb}.
To make this selection, we first excluded datasets where all previously benchmarked algorithms achieved $R^2 > 0.99$, or if a simple linear regression was also capable of achieving this same $R^2$ threshold.
Next, we created a numerical feature vector for each dataset containing the number of samples, number of features, and the performance of each algorithm tested in the SRBench 1.0. We then applied the t-SNE algorithm to reduce this metadata of the datasets to two dimensions and created $12$ clusters using the $k$-means algorithm over the latent encodings. Finally, we picked the dataset closest to each centroid. The final selection of datasets consists of different combinations of horizontal and vertical dimensionality and, as a byproduct, different codomains ---an important factor in proper model fitting, as these datasets may deviate from the traditional assumption of a normal distribution, requiring algorithms to adapt accordingly.

We should also note that in the original SRBench, half of the black-box problems were variations of the Friedman datasets~\cite{Friedman2001}, which introduced bias in the results, as reported by \citet{10.1145/3597312}. To mitigate this issue, we restricted the selection of these datasets, allowing a maximum of $25\%$ of the chosen datasets to belong to this class.
The goal is to ensure that the datasets provide a diverse representation of SR applications while enabling fine-grained analyses to better understand why certain methods outperform others.

The phenomenological \& first-principles track includes publicly available datasets from~\citet{multiviewsr} and \citet{cranmer2023interpretablemachinelearningscience}. The former provides real-world measurements used to derive \emph{empirical relationships}, while the latter source data underlying well-known first-principles equations. All these datasets are particularly challenging because of the small number of points and the diverse and unknown natural distribution of the phenomena.

\subsection{Benchmarked methods}

\begin{table*}[htb]
    \footnotesize
    \caption{
        Algorithms evaluated, their original references, and relevant characteristics pertinent to benchmarking.
    }
    \label{tab:methods_and_hyperparameters}
    \begin{tabular}{@{}r cccc c p{10cm}@{}}
    \toprule\midrule
    \textbf{Algorithm} & \makecell{\textbf{Const.}\\\textbf{Opt.}} & \makecell{\textbf{Time}\\\textbf{limit}} & \makecell{\textbf{Multiple}\\\textbf{solutions}} & \makecell{\textbf{Runs}\\\textbf{on}} & \textbf{Language} & \textbf{Description} \\ \midrule
     
    AFP~\cite{Schmidt2010} & \xmark & \checkmark & \xmark & CPU & \CPP{} & Age-fitness Pareto (AFP) optimization, meaning model age is used as an objective, with constants randomly changed \\ \midrule
    
    AFP\_fe & \xmark & \checkmark & \xmark & CPU  & \CPP{} & AFP with co-evolved fitness estimates \\ \midrule

    AFP\_ehc~\cite{la2016inference} & \checkmark & \checkmark & \xmark & CPU  & \CPP{} & AFP with epigenetic hill climbing for constants optimization as local search  \\ \midrule
    
    Bingo~\cite{randall2022bingo} & \checkmark & \checkmark & PF & CPU  & Python & Evolves acyclic graphs with non-linear optimization, using islands for managing parallel populations\\ \midrule

    Brush~\cite{brush_multi_armed_bandits} & \checkmark & \checkmark & PF & CPU & \CPP{} & GP with multi-armed bandits for controlling search space exploration\\ \midrule
    
    BSR~\cite{jin2020bayesiansymbolicregression} & \xmark & \checkmark & \xmark & CPU & Python & Bayesian model with priors for operators and coefficients is used to sample expression trees \\ \midrule
    
    E2E~\cite{kamienny2022endtoend} & \xmark & \xmark & \xmark & GPU & Python & Generator using pre-trained transformers, using BFGS and subsampling for tuning parameters\\ \midrule

    EPLEX~\cite{eplex} & \xmark & \checkmark & \xmark & CPU  & \CPP{} & GP with $\epsilon$-lexicase parent selection \\ \midrule

    EQL~\cite{pmlrv80sahoo18a} & \xmark & \xmark & \xmark & CPU  & Python & Shallow neural network using mathematical operators as activation functions, and performs a pruning to refine the network to an expression \\ \midrule
    
    FEAT~\cite{cava2018learning} & \checkmark & \checkmark & PF & CPU & \CPP{} & GP algorithm with $\epsilon$-lexicase selection and linear combination of expressions using L1-OLS\\ \midrule
    
    FFX~\cite{mcconaghy2011ffx} & \checkmark & \xmark & \xmark & CPU & Python & Non-evolutionary, deterministic approach, that generates a set of base functions and fits a regularized OLS to combine them\\ \midrule
    
    \makecell{Genetic\\Engine~\cite{espada2022data}} & \xmark & \checkmark & \checkmark & CPU & Python & GP using Context-Free Grammars to guide the generation process in an efficient manner \\ \midrule

    GPGomea~\cite{Virgolin_2021}  & \checkmark & \xmark & \checkmark & CPU & \CPP{} & GP with linkage learning used to propagate patterns and avoid their disruption\\ \midrule
    
    GPlearn  & \xmark & \xmark & \checkmark & CPU  & Python & Canonical GP implementation \\ \midrule

    GPZGD~\cite{10.1145/3377930.3390237}  & \checkmark & \checkmark & \xmark & CPU & C & GP with Z-score standardization and stochastic gradient descent for parameter optimization \\ \midrule
    
    ITEA~\cite{itea}  & \checkmark & \xmark & \xmark & CPU  & Haskell & Mutation-based algorithm with constrained representation and OLS parameter optimization\\ \midrule
    
    NeSymRes~\cite{pmlr-v139-biggio21a} & \checkmark & \checkmark & \xmark & GPU  & Python & Pre-trained encoder-decoders generate equation skeletons, optimized with non-linear optimization\\ \midrule
    
    Operon~\cite{kommenda2020parameter} & \checkmark & \checkmark & PF & CPU  & \CPP{} & GP algorithm with weighted terminals and non-linear OLS parameter optimization \\ \midrule

    Ps-Tree~\cite{pstree} & \xmark & \xmark & \xmark & CPU  & Python & GP algorithm that evolves Piecewise trees with SR expressions as leaves\\ \midrule

    PySR~\cite{cranmer2023interpretablemachinelearningscience} & \checkmark & \checkmark & \xmark & CPU  & Julia & Evolve-simplify-optimize loop with islands to manage parallel populations \\ \midrule

    Qlattice~\cite{broløs2021approachsymbolicregressionusing} & \checkmark & \checkmark & \xmark & CPU  & Python & Uses a learned probability distribution updated over iterations to sample expressions, with parameter optimization. Closed source software \\ \midrule
    
    Rils-rols~\cite{Kartelj2023} & \checkmark & \checkmark & \xmark & CPU  & \CPP{} & Iterative generation of perturbations and parameter optimization with OLS and local search selection of next candidates\\ \midrule
    
    TIR~\cite{tir} & \checkmark & \checkmark & PF & CPU & Haskell & GP with crossover and mutation that uses a constrained representation capable of tuning non-linear parameters with OLS \\ \midrule

    TPSR~\cite{shojaee2023transformerbased} & \checkmark & \checkmark & \xmark & GPU  & Python & Monte-Carlo Tree Search planning with non-linear optimization wrapper for generative models E2E and NeSymRes\\ \midrule
    
    uDSR~\cite{NEURIPS2022_dbca58f3} & \checkmark & \xmark & \xmark & GPU  & Python & Unification of pre-trained transformers, GP, and linear models, into a framework that decomposes the problem \\ 
    
    \midrule\bottomrule
    \end{tabular}
\end{table*}

In this benchmark, we have extended the list of methods from the original SRBench adding recently published methods, reported in \cref{tab:methods_and_hyperparameters}, highlighting whether the algorithm supports parameter optimization, a time-limit argument\footnote{In case it does not, we adjusted the hyperparameters to reasonable values to stop at an approximate time limit.}, whether it returns multiple solutions (\textit{e.g.} a Pareto Front (PF)) or just a single solution, the hardware the algorithm runs on, whether grid search to finetune the hyper-parameters is possible, and the programming language used in the implementation.
For methods implemented in other programming languages, the authors also implemented Python bindings to facilitate their use.

A grid search was performed for all methods, except for three ---Bingo, Brush, and GP-GOMEA--- due to incompatibility with scikit-learn's \texttt{GridSearchCV} interface at the time. We adopted the grid search settings from the original SRBench and defined a new search space for methods that lacked one. An exception is FFX, which has no tunable hyperparameters.
Additionally, we included the default (off-the-shelf) configuration as one of the candidate settings, since preliminary results suggest that, for some problems, is outperforms tuned configurations.
For further details, refer to the online resources in \cref{apx:online_resources}.

Hyper-parameter optimization is performed before each run using a $3$-fold cross-validation on the training data ($75\%$) and selecting the best configuration based on the average $R^2$ score. The model is then trained with optimal hyperparameters using the entire training data, and evaluated on a held-out test data (25\%).
We implemented a \emph{Python} wrapper to tune every method into the same grid search pipeline using \textit{scikit-learn}.

\subsection{Performance analysis}

We repeated these experiments $30$ times for each dataset with fixed and distinct seeds, storing all necessary information to process the aggregated results.
For SRBench 2.0, we adopted the \textit{performance profile plot}~\cite{dolan2002benchmarking}, which describes the empirical distribution of the obtained results. This plot illustrates the probability of achieving a performance greater than or equal to a given $R^2$ threshold for all possible thresholds. In this plot, the $x$-axis represents a threshold value of the $R^2$ and the $y$-axis the percentage of runs that a particular algorithm obtained that value or higher. This plot can give a broader view of the likelihood of successfully achieving a high accuracy for each algorithm, while keeping all the information about each algorithm in the same plot. Using this same plot, we can calculate the area under the curve (AUC) for each individual algorithm as an aggregated measure: a value of $1.0$ indicates that all $30$ runs achieved the maximum $R^2=1$, whereas a value of $0.0$ indicates that all runs resulted in $R^2 \leq 0$. The AUC provides an estimate of the method’s potential to achieve high performance across multiple runs.
The performance plot can be generated for each individual dataset as a result of the $30$ independent runs or we can plot the aggregated performance over all datasets using an aggregation function (\textit{e.g.}, mean, median, max, min) for each dataset. In this paper, we use the maximum value as the aggregation function, which reflects the best-case performance of each algorithm across the $30$ runs. The project website reports both individual and aggregated plots.

To measure model size, we convert the models to a SymPy~\cite{10.7717/peerj-cs.103} compatible expression and count the number of nodes. This is necessary since the internal complexity measures of each algorithm may differ from the standard way of counting nodes.
The expression was not simplified (except for trivial simplifications such as constant merging) before counting the nodes, as simplification using SymPy is unreliable and can increase the model size in the process~\cite{10.1145/3583131.3590346}.

\subsection{Benchmark infrastructure}

Conducting extensive experiments across multiple algorithms and datasets requires substantial computational resources. Variability in hardware workload or non-homogeneous cluster nodes can make it challenging to establish a fixed computational budget or ensure that all methods are evaluated under identical conditions. Using fixed resources can help mitigate the "hardware lottery" effect~\cite{hooker2020hardwarelottery}, where an approach succeeds due to its compatibility with the available software and hardware rather than its inherent merit.

Experiments were executed on a single computing cluster. Each job was allocated $10$ GB of RAM, and GPUs were provided for methods that support them (see the \textit{runs on} column in \Cref{tab:methods_and_hyperparameters}). Jobs were not allowed to spawn subprocesses or utilize multiple cores. The time budget was set to $6$ hours for hyperparameter search and $1$ hour for training the algorithm with the optimal configuration, after which a \texttt{SIGALRM} signal was sent to terminate the process.

Energy consumption was estimated after hyperparameter tuning to avoid additional overhead, using the \textit{eco2AI} library~\cite{Budennyy2022}, which derives estimates from standard Linux commands monitoring CPU and memory usage. While not exact, this provides a rough yet informative profile of resource usage.

\subsection{Data Availability}~\label{apx:online_resources}

The datasets were submitted to PMLB~\cite{Olson2017PMLB, romano2021pmlb}.
We released all experiment scripts and parameters to ensure transparency and reproducibility.
We also restructured SRBench using containerized environments to ensure consistent execution across systems and simplify standalone use.
The project is available at \url{https://github.com/cavalab/srbench/tree/srbench_2025}.

\section{Results}~\label{sec:results}

\begin{figure}[htb]
    \centering
    \includegraphics[width=\linewidth]{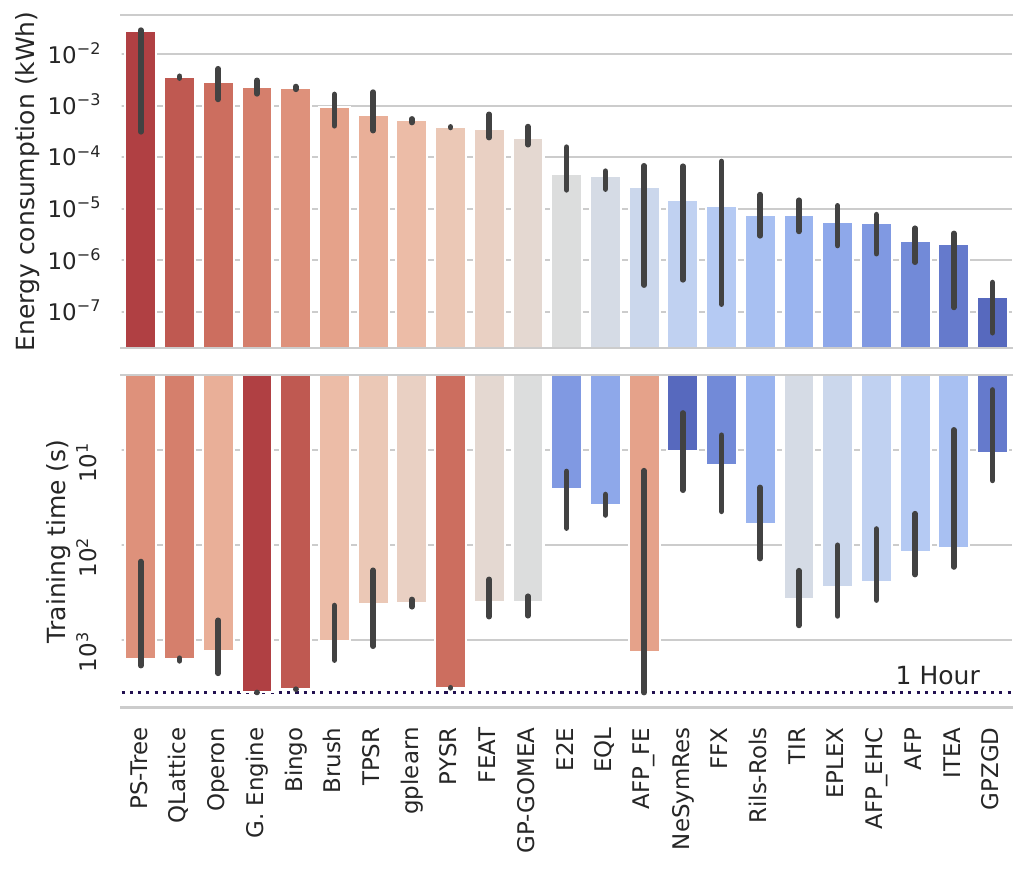}
    \caption{Median energy consumption (kWh) and training runtime for each algorithm.}
    \label{fig:power_consumption}
\end{figure}

Within the specified time budget, the total runtime on a single core would be $1$ year, $43$ weeks, and $5$ days.
Using eco2AI to monitor the final training phase, we report the energy and time consumption in \cref{fig:power_consumption}.
We notice that these measurements are influenced not only by the choice of programming language and implementation details but also by the internal decisions of the algorithms to terminate execution before reaching the maximum time.

\begin{figure*}[htb]
    \centering
    \includegraphics[width=\linewidth]{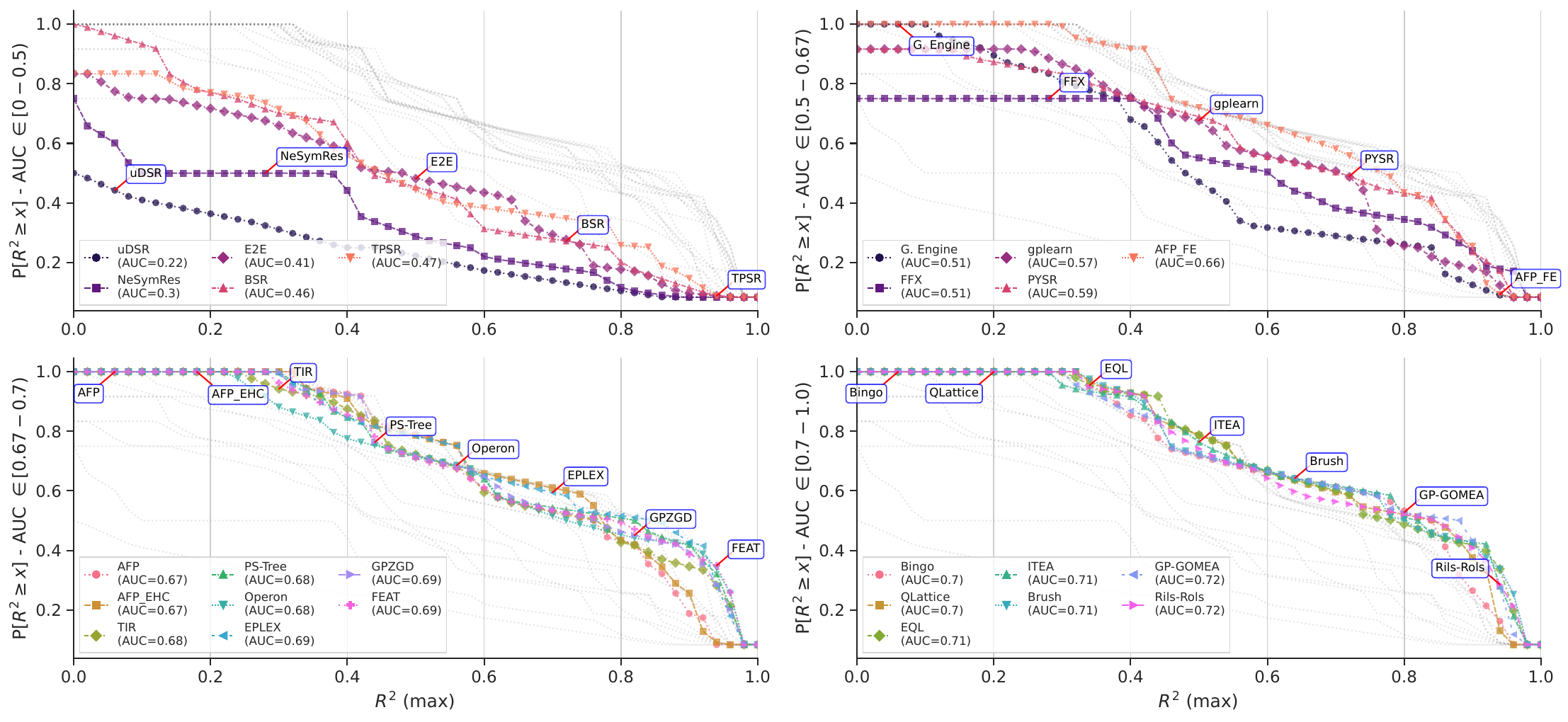}
    \caption{Performance plots for the black-box track, where the lines represent the probability of obtaining a given empirically observed $R^2$ value when running the experiments multiple time (i.e., max aggregation).
    }
    \label{fig:perfplot_max}
\end{figure*}

In \Cref{fig:perfplot_max}, we present the performance plot with the AUC values based on the highest empirically observed $R^2$ on the test set across $30$ runs, representing an optimistic perspective on model performance, assuming a user performs the same number of runs. 
The plot shows the empirically observed likelihood of obtaining a maximum $R^2$ score for black-box problems under sufficient repetitions.
The probability on the $y$-axis, $\text{P}[R^2\geq x]$, is estimated by calculating the proportion of runs where the maximum $R^2$ exceeds a given threshold $x$.

\begin{figure*}[htb]
    \centering
    \includegraphics[width=\linewidth]{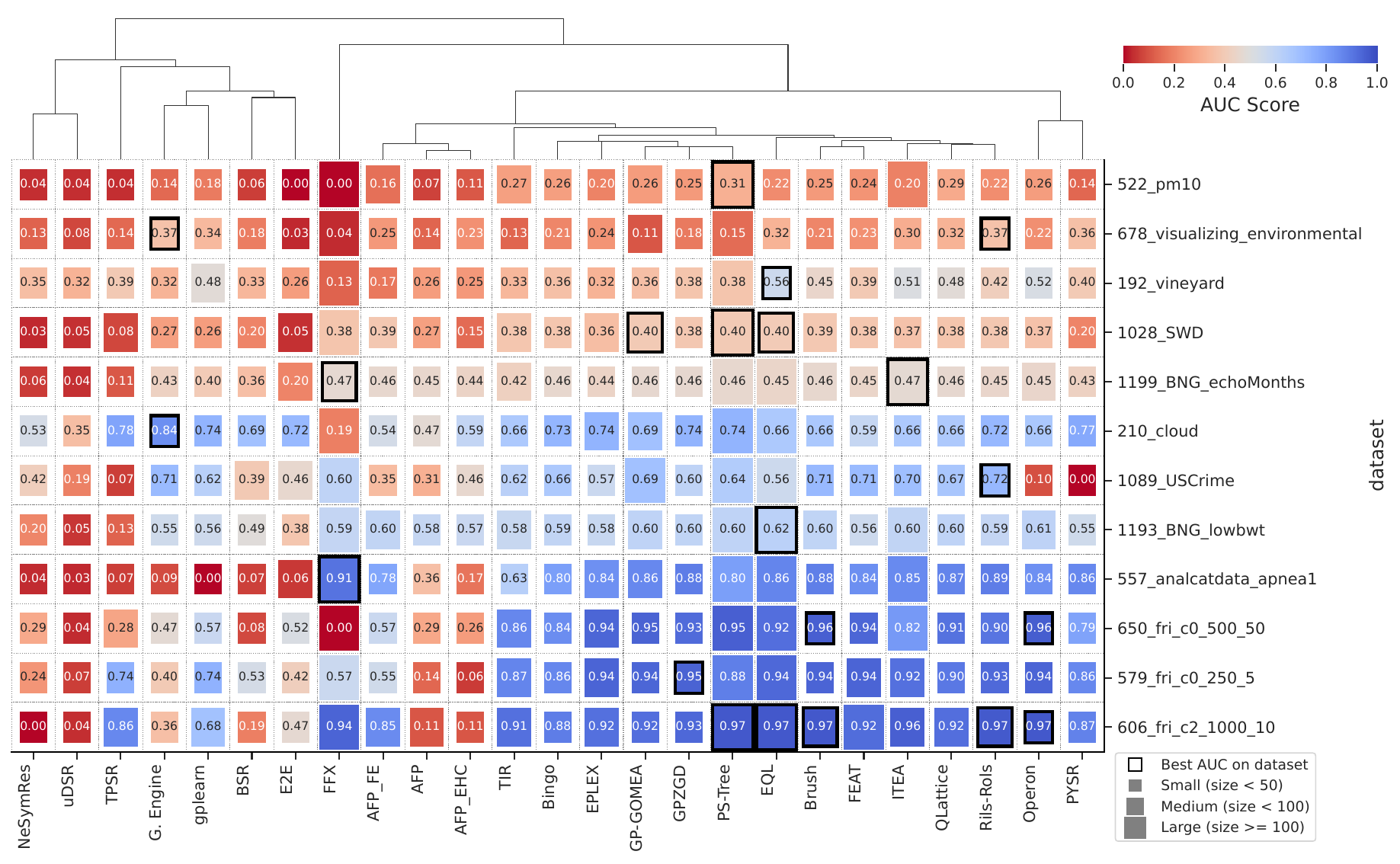}
    \caption{
        Cluster map of the Area Under the Curve (AUC) of Expected Performances across the $30$ independent runs is segregated by algorithm and dataset. Higher values indicate better performance, while larger cells represent worse model size.
    }
    \label{fig:perfplot_clustermap_with_sizes}
\end{figure*}

\Cref{fig:perfplot_clustermap_with_sizes} shows a cluster map of the AUC from the performance plot for each pair of dataset and algorithm on the test set. The size of each cell is proportional to the size of the best-performing final expression across the 30 runs. Higher values and smaller cells indicate better performance. At the top of this plot, we can observe a hierarchical clustering of the algorithms, grouping the methods based on performance, and the datasets based on their difficulty to solve. The best algorithm for each dataset is highlighted with a black edge color on its cell.

\begin{figure*}[htb]
    \centering
    \includegraphics[width=\linewidth]{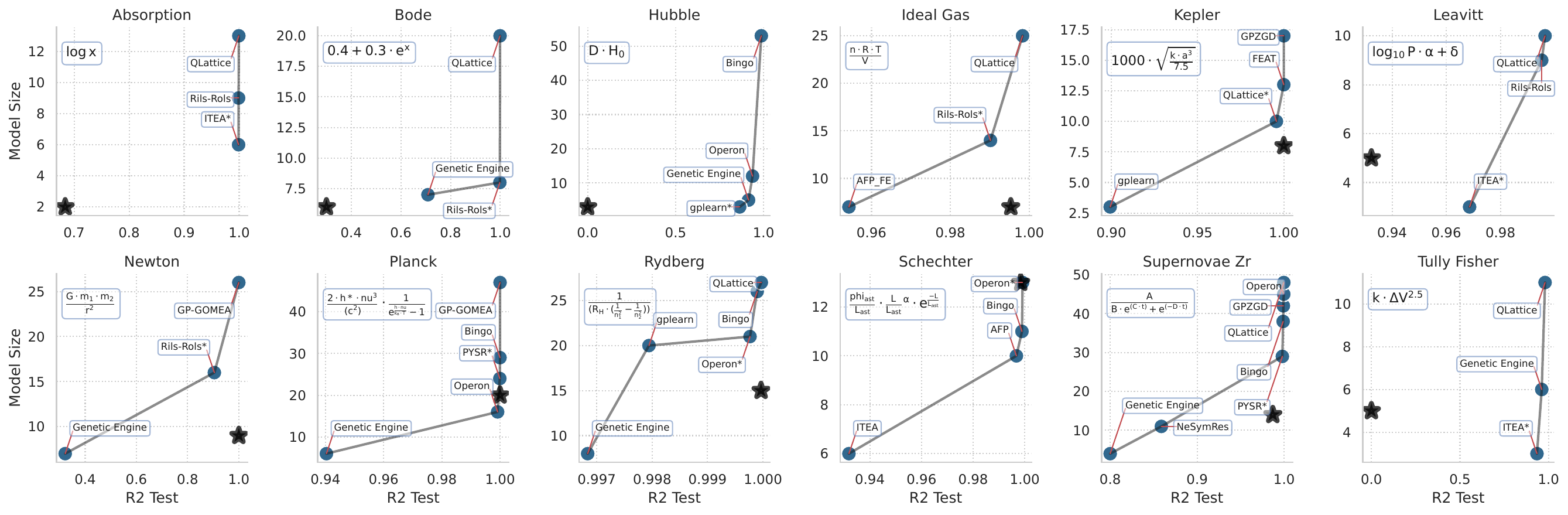}
    \caption{Pareto plots for the phenomenological \& first-principles track, with model sizes on $y$ axis, and $R^2$ on $x$ axis. The \textit{star} marker denotes the ground-truth expression performance, also denoted in the box inside each subplot. Only the first front is plotted for clarity.}
    \label{fig:Pareto_first_principles}
\end{figure*}

For the phenomenological \& first-principles track, \Cref{fig:Pareto_first_principles} displays the Pareto front on the test set of the most accurate solutions obtained by SR algorithms and compared to the current accepted hypothesis (\textit{star}).
We observe that noise ---whether inherent in real-world data or synthetically added in the case of generated data--- can cause the ground-truth hypotheses to fall short of achieving a perfect $R^2$.
Asterisks next to the algorithm names indicate those closest (based on Euclidean distance over the normalized axes) to the ground-truth.
The final equations closest to the governing models are shown in \Cref{tab:Pareto_first_principles_equations} along with their corresponding $R^2$ scores and sizes.

\begin{table}[htb]
\caption{Equation from the Pareto front closest to the governing models.}
\label{tab:Pareto_first_principles_equations}
\footnotesize
\centering
\begin{tabular}{@{}rp{0.25cm}p{0.25cm}p{5.475cm}@{}}
    \toprule\midrule
    \textbf{Dataset} &  $R^2$ & \textbf{Size} & \textbf{Symbolic model} \\ \midrule
    absorption & $1.0$ & $6$ & $0.24 + 1.76\tanh{x}$ \\ \midrule
    bode & $1.0$ & $8$ & $0.34 e^{1.51\cdot n} - 0.87$ \\ \midrule
    hubble & $0.86$ & $3$ & $0.090+D$ \\ \midrule
    ideal\_gas & $0.99$ & $14$ & $0.69\cdot \log{n + 1.64} - 0.89 + 0.48*e^{-V}$ \\ \midrule
    kepler & $1.00$ & $10$ & $1.98*\tanh{(0.66\cdot a - 0.56)} + 0.78$ \\ \midrule
    leavitt &$ 0.97$ & $3$ & $-0.94 \cdot \log{P}$ \\ \midrule
    newton & $0.90$ & $16$ & $0.34\cdot \frac{m_1}{0.83\cdot m_2 - \frac{0.09}{m_1}} + 1.13$ \\ \midrule
    planck & $1.00$ & $24$ & $0.023\cdot \log{(\sqrt{nu + 0.38} - 0.04)} - 0.31 \cdot \frac{nu + 0.3}{T + 0.94} + 0.41$ \\ \midrule
    rydberg & $1.00$ & $21$ & $1.01\cdot e^{0.1 \cdot e^{1.73 \cdot n_1 -1.31 \cdot n_2}} - e^{-0.69 \cdot n_1}$ \\ \midrule
    schechter & $1.00$ & $13$ & $0.748 - 0.274 \cdot (\log{(163.992 \cdot L + 106.737)} + 1.414 \cdot L)$ \\ \midrule
    supernovae\_zr & $1.00$ & $29$ & $(\sin{(0.2\cdot x\cdot e^{-x})} - 0.04) \cdot (x + 0.72\cdot \sin{(5.46 \cdot x + 0.76)} - 0.16) - \sin{(x + 0.18)}$ \\ \midrule
    tully\_fisher & $0.93$ & $3$ & $-0.93\cdot \Delta V$ \\ \midrule\bottomrule
\end{tabular}
\end{table}

\section{Discussion}~\label{sec:discussions}

\subsection{Energy consumption and execution time}

Energy consumption profiles in \Cref{fig:power_consumption} reveal distinct patterns for different algorithms.

The top two most energy-hungry algorithms are Python-based, and the third, Operon, is implemented in \CPP{} but includes an inner hyperparameter tuning step. Some methods, such as FEAT and Brush, implement a \texttt{max\_stall} parameter, which halts execution if no improvement is observed in an inner validation partition after a number of iterations. While this feature does not necessarily correlate with these methods being the best performers in terms of energy consumption, it could be further explored for more computationally demanding algorithms.

GPU-based algorithms, such as TPSR, E2E, and NeSymRes, do not show much difference in energy consumption compared to CPU-based algorithms. In particular, TPSR, which uses traditional parameter optimization methods, demonstrated higher execution times than other GPU-based methods.
uDSR was not compatible with the library during the experiments.

\subsection{Performance analysis of the \textit{black-box} track}

The performance curves in \Cref{fig:perfplot_max} displays the expected probability of achieving maximum performance at different $R^2$ thresholds across a variety of problems.
Some methods (shown in the top-left subplot, with AUC below $0.5$) exhibit a probability lower than $1.0$ of achieving a strictly positive $R^2$, suggesting that they may struggle to generalize; fail to capture intrinsic data structures, or produce low-quality models.

Most of the top-performing algorithms incorporate constant optimization as a local search step ---using linear, gradient-based, or non-linear methods--, highlighting the importance of parameter optimization for achieving high performance.
Among the highest AUC scores are GP-based algorithms such as Brush, GP-GOMEA, and GPZGD. Notably, GPZGD is a GP algorithm that standardizes the features using Z-score and performs parameter optimization.
Other methods, such as Rils-Rols, also performs constant optimization. 

The selection of datasets in \Cref{fig:perfplot_clustermap_with_sizes} reveals the ``\textit{no free lunch}'' aspect of the current state of SR --- there is no algorithm that performs exceptionally well on all datasets.
Likewise, there is no algorithm that fails to solve all problems. Hierarchical clustering shows two main clusters of good- and bad-performing algorithms based on the AUC of the performance curves.
A closer inspection of \Cref{fig:perfplot_clustermap_with_sizes} allows for a better understanding of challenges for existing algorithms.

Furthermore, the blue cells reveals a cluster of datasets in which the top-performing algorithms obtain a similarly good result. At the same time, we can readily see the trade-off between performance AUC and size. For example, in the \emph{557} dataset, we can see that FFX obtained the best AUC score, while Rils-Rols obtained a slightly worse AUC result but with a smaller expression.
Interestingly, some methods that perform poorly overall still achieve the best results on specific datasets. This highlights that no single algorithm dominates across all objectives; each has its own strengths depending on the problem characteristics.

\subsection{Towards a parameter-less experience}

Setting the hyper-parameters of SR algorithms often requires careful experimentation with a number of different settings that may affect the final result.
For general users, this creates a significant barrier: they must either rely on default values or conduct extensive experimentation to find the best configurations. The field should move toward a parameter-less experience by developing algorithms that require as few hyperparameters as possible, unless they are intuitive and domain-related.
Good practices involve carefully selecting default parameters and, if possible, integrating internal hyperparameter tuning (such as Operon with Optuna) or making parameters adaptive (as in PySR).

We propose standardizing the hyperparameter tuning by requiring each method to define a small set (\textit{e.g.}, four) of hyperparameter configurations.
A simple grid search is then applied using these configurations, each limited to a maximum runtime (\textit{e.g.}, one hour).
The off-the-shelf configuration should automatically be included by the experiment scripts, as we observed some methods perform best with their default settings.

\subsection{Solving empirical and theoretical problems}

The phenomenological \& first-principles track provides a detailed view of the trade-off between model size and accuracy, using established models as reference points.
\cref{fig:Pareto_first_principles} shows that, for almost every problem, no SR method could discover an expression that strictly dominates the original equation. This occurs because methods either generate a larger expression with higher accuracy or a smaller expression with lower accuracy.
\cref{tab:Pareto_first_principles_equations} further illustrates this challenge --- only ITEA successfully retrieved one of the first-principles equations (Leavitt). In other cases, SR methods either produced a more complex yet more accurate expression or a simpler but less precise model. For the \emph{Schechter} dataset, Operon found reasonable alternatives without significantly deviating from the complexity-accuracy balance of the original equation.

We notice that some first-principle equations fails to achieve perfect $R^2$ due to noise in the original data --- notably in the Hubble and Tully-Fisher datasets.
In such cases, obtaining a higher $R^2$ with a more complex model often indicates that the model is fitting the noise rather than the underlying relationship. This is evident in the equations found for Hubble and Tully-Fisher, which deviate from the ground-truth equation by only a small edit distance.

In contrast, the Absorption and Bode datasets represent phenomenological problems with no known ground-truth equations.
A possible venue for future investigation, is the incorporation of the uncertainties information of the data, leading to a better measurement for the model accuracy.

\subsection{Current difficulties and limitations}

During the benchmarking process, a number of bugs had to be fixed for most evaluated algorithms. 
Few of the corresponding repositories are being maintained, requiring a joint decision of which algorithm to deprecate and remove from further benchmarking. 

We propose establishing deprecation rules to maximize benchmarking efforts. A simple criterion could be: if an algorithm’s repository is no longer actively maintained and it does not appear on at least one of the Pareto fronts in the black-box datasets, it should be deprecated. 
This guideline may also encourage the development of more robust algorithms that perform consistently well across diverse problems.

Another important point is the fair comparison between GPU-based and CPU-based approaches, as well as single-core versus multi-core implementations. While every algorithm with a \texttt{max\_time} hyperparameter respects its limit, some terminate far earlier (\textit{e.g.}, NeSymRes), potentially skewing performance evaluations. A more precise implementation of time management would allow for a fairer comparison of computational efficiency across different methods.

Additionally, this benchmark could provide significantly more data-driven insights for algorithm development by incorporating appropriate metadata (\textit{e.g.}, a set of keywords), offering a direct way to link methodologies to results.
Similarly, algorithms could generate post-run metadata (\textit{e.g.}, results of internal grid search, number of local search iterations) as part of their output. This would facilitate a deeper understanding of algorithmic behavior and enable more precise fine-tuning of specific methods, ultimately leading to more informed improvements in SR techniques.

\section{Conclusions and future work}~\label{sec:conclusions}

This paper updates the current SRBench aiming to improve the benchmark of symbolic regression as a community accepted standard. We increased to almost twice the number of SR algorithms than the previous version, and provided an alternative and more detailed visualizations depicting the current state of SR. 

To ensure a constant improvement of SRBench, we need active participation of the community to ensure compatibility with the current experiments and a correctly working implementation. We \emph{call for action} from researchers to actively contribute to discussions, share new ideas, provide constructive criticism, and propose corrections. A continuously evolving benchmark will help drive progress in SR.
Achieving a mature state for the field should be a shared priority, especially given the growing emphasis on transparency in machine learning and the increasing interest in scientific ML.

\begin{acks}
    To everyone involved in the discussions to improve SRBench. To PMLB maintainers for the help. To all past researchers who somehow enabled us to use data or methods from their experiments, making science open.
    
    Alcides Fonseca is supported by FCT through the LASIGE Research Unit, ref. UID/000408/2025.
    Fabricio Olivetti de Franca is supported by Conselho Nacional de Desenvolvimento Cient\'{i}fico e Tecnol\'{o}gico (CNPq) grant 301596/2022-0.
    Guilherme Imai Aldeia is supported by Coordena\c{c}\~{a}o de Aperfei\c{c}oamento de Pessoal de N\'{i}vel Superior (CAPES) finance Code 001 and grant 88887.802848/2023-00.
\end{acks}

\bibliographystyle{ACM-Reference-Format}
\bibliography{refs}

\end{document}